\documentclass[10pt,twocolumn,letterpaper]{article}

\usepackage{cvpr}
\usepackage{times}
\usepackage{epsfig}
\usepackage{graphicx}
\usepackage{amsmath}
\usepackage{amssymb}
\usepackage{float}
\usepackage{subfigure}
\usepackage{multirow}

\usepackage[pagebackref=true,breaklinks=true,letterpaper=true,colorlinks,bookmarks=false]{hyperref}

\cvprfinalcopy 

\def\cvprPaperID{1497} 

\begin{document}

\title{CrowdPose: Efficient Crowded Scenes Pose Estimation and A New Benchmark}

\author{Jiefeng Li$^{1*}$, Can Wang$^1$, Hao Zhu$^1$, Yihuan Mao$^2$, Hao-Shu Fang$^1$, Cewu Lu$^{1\dag}$\\
$^1${Shanghai Jiao Tong University}, $^2${Tsinghua University}\\
{\tt\small \{{ljf\_likit},{wangcan123},{lucewu}\}@{sjtu.edu.cn}}\\
{\tt\small {haozhu}@{zju.edu.cn}} {\tt\small {maoyh16}@{mails.tsinghua.edu.cn}} {\tt\small {fhaoshu}@{gmail.com}}
}

\maketitle

\begin{abstract}
  Multi-person pose estimation is fundamental to many computer vision tasks and has made significant progress in recent years. However, few previous methods explored the problem of pose estimation in crowded scenes while it remains challenging and inevitable in many scenarios. Moreover, current benchmarks cannot provide an appropriate evaluation for such cases. In this paper, we propose a novel and efficient method to tackle the problem of pose estimation in the crowd and a new dataset to better evaluate algorithms. Our model consists of two key components: joint-candidate single person pose estimation (SPPE) and global maximum joints association. With multi-peak prediction for each joint and global association using graph model, our method is robust to inevitable interference in crowded scenes and very efficient in inference.  The proposed method surpasses the state-of-the-art methods on CrowdPose dataset by \textbf{5.2} mAP and results on MSCOCO dataset demonstrate the generalization ability of our method. Source code and dataset will be made publicly available.
\end{abstract}

\section{Introduction}
Estimating multi-person poses in images plays an important role in the area of computer vision. It has attracted tremendous interest for its wide applications in activity understanding~\cite{activity1, activity2}, human re-identification~\cite{reid1}, human parsing~\cite{parsing1, parsing2} etc. Currently, most of the methods can be roughly divided into two categories: i) top-down approaches that firstly detect each person and then perform single person pose estimation, or ii) bottom-up approaches which detect each joint and then associate them into a whole person.

\begin{figure}[tb]
\begin{center}
\includegraphics[width=0.8\linewidth]{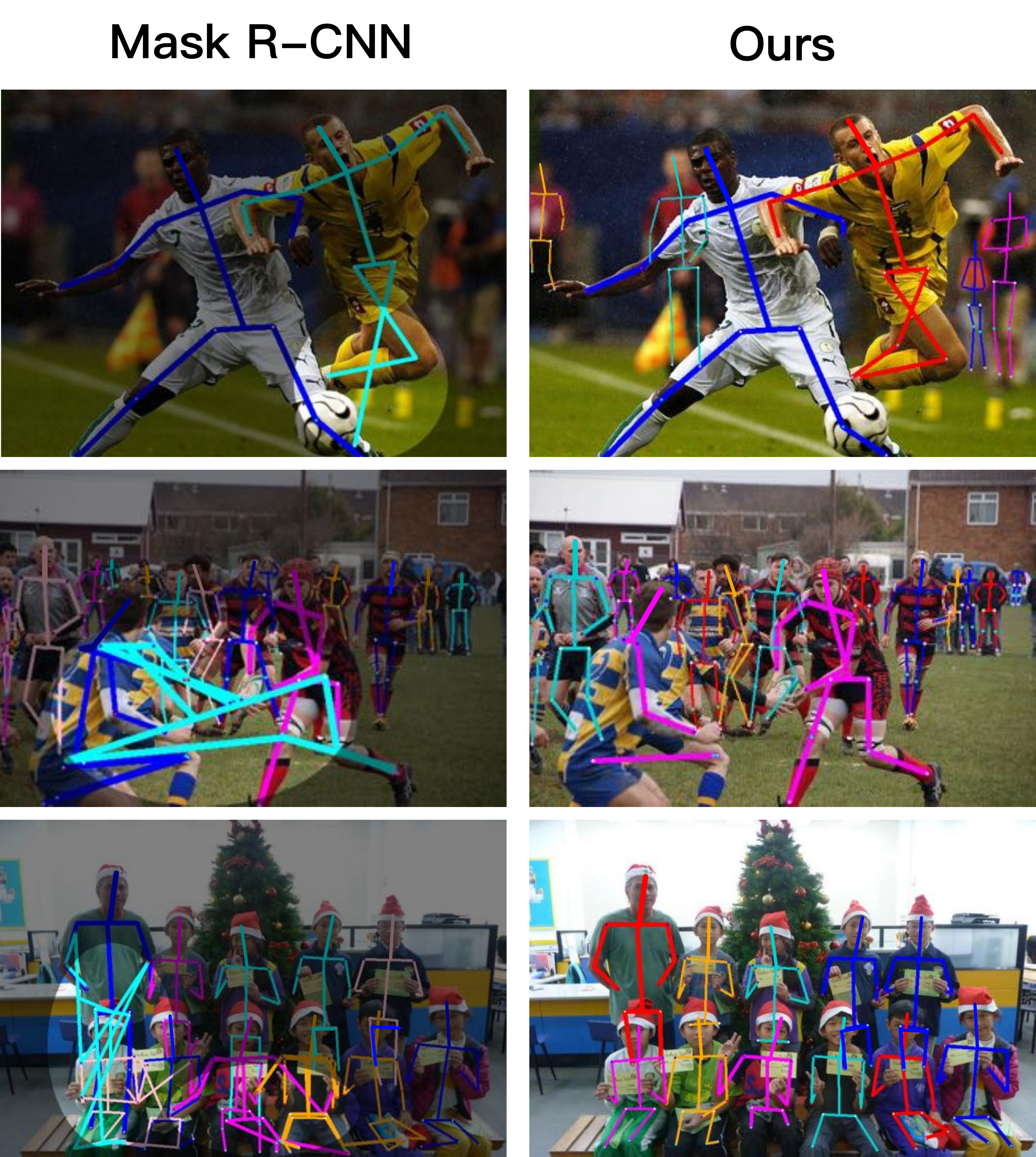}
\end{center}
   \caption{Qualitative comparison of Mask R-CNN and our CrowdPose method in crowded scenes. Though current methods achieve good performance in public benchmarks~\cite{mpii, mscoco, aic}, they fail in crowded cases. There are mainly two types of errors: i) assemble wrong joints into a pose; ii) predict redundant poses in crowded scenes.}
\label{fig:diff}
\vspace{-0.15in}
\end{figure}

\begin{figure*}[!t]
\begin{center}
\includegraphics[width=0.9\linewidth]{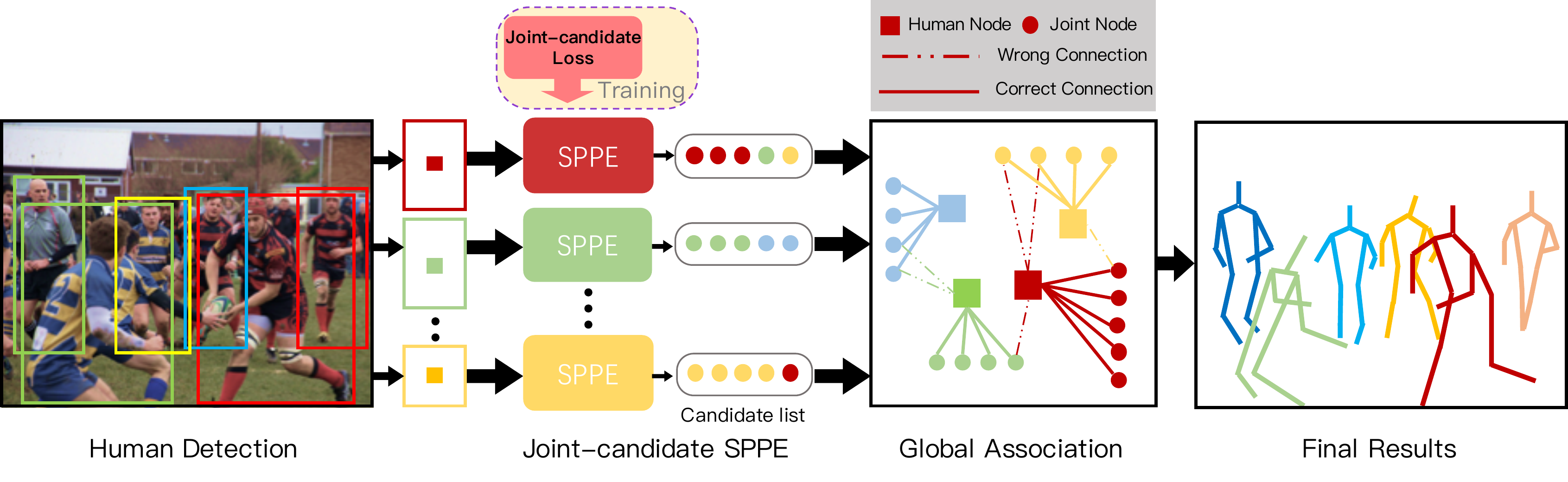}
\end{center}
   \caption{Pipeline of our proposed method. JC SPPE uses joint-candidate loss function during training phase. In inference phase, JC SPPE receives human proposals and generates joint candidates. Then we utilize human proposals and joint candidates to build a person-joint graph. Finally, we associate joints with human proposals by solving the assignment problem in our graph model.}
\label{fig:Pipeline}
\vspace{-0.15in}
\end{figure*}
\begin{figure}[tb]
\begin{center}
\includegraphics[width=0.8\linewidth]{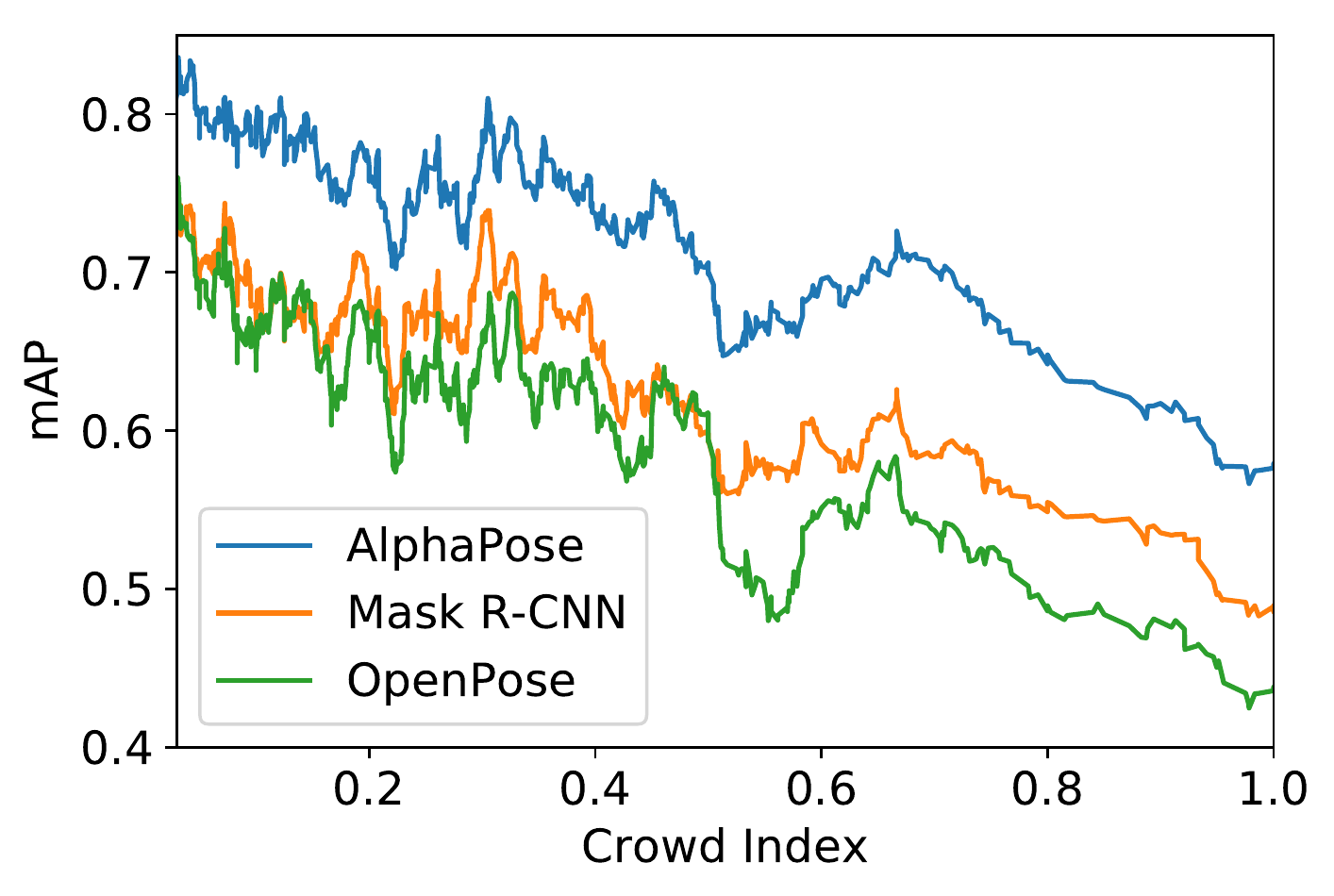}
\end{center}
   \caption{State-of-the-art methods evaluation results on MSCOCO dataset. The x-axis is the \textit{Crowd Index} which we defined to measure the crowding level of an image. Compared to uncrowded scenes, the accuracy (mAP$@$0.5:0.95) of state-of-the-art methods are about 20 mAP lower in crowded cases.}
\label{fig:map}
\vspace{-0.15in}
\end{figure} 

To evaluate the performance of multi-person pose estimation algorithms, several public benchmarks were established, such as MSCOCO~\cite{mscoco}, MPII~\cite{mpii} and AI Challenger~\cite{aic}. In these benchmarks, the images are usually collected from daily life where crowded scenes appear less frequently. As a result, most of the images in these benchmarks have few mutual occlusion among humans. For example, in MSCOCO dataset (persons subset), 67.01\% of the images have no overlapped person. Current methods have obtained encouraging success on these datasets.

However, despite the good performance that current methods have achieved on previous benchmarks, we observe an obvious degradation of their performance in crowded cases. As shown in Figure~\ref{fig:diff}, for the current state-of-the-art methods~\cite{msra, alphapose, maskrcnn, cao} of both bottom-up and top-down approaches, their performance decreases dramatically as the crowd level increases (as Figure~\ref{fig:map}). Few previous methods aimed to tackle the problem of pose estimation in crowded scene, and no public benchmark has been built for this purpose. Meanwhile, crowded scenes are inevitable in many scenarios.

In this paper, we propose a novel method to tackle the problem of pose estimation in a crowd, using a global view to address interference problem. Our method follows the top-down framework, which first detects individual persons and then performs single person pose estimation (SPPE). We propose a joint-candidate SPPE and a global maximum joints association algorithm. Different from previous methods that only predict target joints for input human proposals, our joint-candidate SPPE outputs a list of candidate locations for each joint. The candidate list includes target and interference joints. Then our association algorithm utilizes these candidates to build a person-joint connection graph. At last, we solve the joint association problem in this graph model with a global maximum joints association algorithm. Moreover, the computational complexity of our graph optimization algorithm is the same as the conventional NMS algorithm. 

To better evaluate human pose estimation algorithms in crowded scenes and promote the development in this area, we collect a dataset of crowded human poses. We define a \textit{Crowd Index} to measure the crowding level of an image. Images in our dataset have a uniform distribution of \textit{Crowd Index} among $[0, 1]$, which means only an algorithm that performs well on both uncrowded and crowded scenes can achieve a high score in our dataset.

To sum up, the contributions of this paper are as follows: i) we propose a novel method to tackle the crowded problem of pose estimation; ii) we collect a new dataset of crowded human poses to better evaluate algorithms in crowded scenes. 
We conduct experiments on our proposed method. When using a same ResNet-101 based network backbone, our method surpasses all the state-of-the-art methods by \textbf{5.2} mAP on our dataset. Moreover, we replace the SPPE and post-processing steps in the state-of-the-art method with our module and brings 0.8 mAP improvement on MSCOCO dataset. That is, our method can generally work in non-crowded scenes.  

\vspace{-0.1in}
\section{Related Work}
\subsection{{2D} Pose Estimation Dataset}
Pioneer works on 2D human pose estimation dataset on RGB images involve LSP~\cite{lsp}, FashionPose~\cite{fashionpose}, PASCAL Person Layout~\cite{pascal}, J-HMDB~\cite{jhmdb}, etc. These datasets have contributed to encouraging progress of human pose estimation. However, they only evaluate for single person pose estimation. With the improvement of algorithms, more researchers focus on multi-person pose estimation problems, and several datasets are established, \eg, MPII~\cite{mpii}, MSCOCO~\cite{mscoco}, AI Challenger~\cite{aic}. In spite of the prevalence of these datasets, they suffer from a low-density problem, which makes the current model overfitted to uncrowded scenes. The performance of the state-of-the-art methods decreases as the number of human increases.

\subsection{Multi-Person Pose Estimation}
\paragraph{Part-Based Framework}
Representative works on the part-based framework~\cite{cao, papandreou2018personlab, kocabas2018multiposenet} are reviewed. Part-based methods detect joints and associate them into a whole person. The state-of-the-art part-based methods are mainly different on their association methods. Cao \etal~\cite{cao} associate joints with a part affinity field and greedy algorithm. Papandreou \etal~\cite{papandreou2018personlab} detect individual joints and predict relative displacements for association. Kocabas \etal~\cite{kocabas2018multiposenet} propose a multi-task model and assign joints to detected persons by a pose residual network. The joint detectors in part-based approached are relatively vulnerable because they only consider small local regions and output smaller response heatmaps. 

\vspace{-0.1in}
\paragraph{Two-Step Framework} Our work follows the two-step approach. A two-step approach first detects human proposals~\cite{fasterrcnn, megdet} and then performs single person pose estimation~\cite{hourglass, pyranet}. The state-of-the-art two-step methods~\cite{cpn, msra, alphapose, maskrcnn} achieve significantly higher scores than the part-based methods. However, the two-step approaches highly depend on human detection results, and it fails in crowded scenes~\cite{crowdhuman}. When people stay close to each other in a crowd, it is improbable to crop a bounding box that only contains one person. Some works~\cite{tracking1, tracking2} aim at human tracking in the crowd. As a supplement to them, we propose a novel and efficient method that significantly increases pose estimation performance in crowded scenes, which is robust to human detection results.

\section{Our Method}
The pipeline of our proposed method is illustrated in Figure~\ref{fig:Pipeline}. Human bounding box proposals obtained by human detector are fed into \textit{joint-candidate} (JC) single person pose estimator (SPPE). JC SPPE locates the joint candidates with different response scores on the heatmap (Sec.~\ref{sec:JC_SPPE}). Then our joint association algorithm takes these results and builds a person-joint connection graph (Sec.~\ref{sec:graph}). Finally, we solve the graph matching problem to find the best joint association result with a global maximum joints association algorithm (Sec. \ref{sec:optimize}).

\subsection{Joint-Candidates SPPE} \label{sec:JC_SPPE}
Joint-candidate SPPE receives a human proposal image and outputs a group of heatmaps to indicate human joint locations. Though a human proposal should indicates only one human instance, in the crowded scenarios, we inevitably need to handle a large number of joints from other human instances. Previous works~\cite{msra, cpn, alphapose} use SPPE to suppress interference joints. However, SPPE fails in crowded scenes because their receptive fields are limited by the input human proposals. To address this problem, we propose joint-candidate SPPE with a novel loss designed in a more global view.

\vspace{-0.1in}
\subsubsection{Loss Design}
For the $i^{th}$ human proposal, we input its region $\mathbf{R}_i$ into our SPPE network and get the output heatmap $\mathbf{P}_i$. There are two types of joints in $\mathbf{R}_i$, that is, the joints belong to the $i^{th}$ person, and the joints belong to other human instances (not the $i^{th}$ person). We name them as target joints and interference joints respectively. 

Our goal is to enhance target joints response and suppress interference joints response. However, we don't suppress them directly since interference joints for the current proposal can be regarded as target joints for other proposals. Thus, we can leverage interference joints to estimate human poses with other human proposals in a global manner. Therefore, to utilize those two kinds of joint candidates, we output them with different intensities. 

\vspace{-0.1in}
\paragraph{Heatmap Loss} For the $k^{th}$ joint in the $i^{th}$ person, we denote the target joint heatmap as $\mathbf{T}_i^k$, consisting of a 2D Gaussian $G(\mathbf{p}_i^k | \sigma)$, centered at the target joint location $\mathbf{p}_i^k$, with standard deviation $\sigma$.

For interference joints, we denote them as a set $\Omega_i^k$. The heatmap of interference joints is denoted as $\mathbf{C}_i^k$, consisting of a Gaussian mixture distribution $\sum_{p \in \Omega_i^k} G(\mathbf{p} | \sigma)$.

Our proposed loss is defined as,
\begin{equation}
    \mathit{Loss}_i = \frac{1}{K} \sum_{k=1}^K MSE[\textbf{P}_i^k, \textbf{T}_i^k + \mu\textbf{C}_i^k]
\end{equation}
where $\mu$ is an attenuation factor ranged in [0,1]. As aforementioned, interference joints will be useful in indicating joints of other human instances. Therefore, we should consider it in a global view by cross-validation. Finally, we have $\mu = 0.5$, which fits our intuition: interference joints should be attenuated but not over-suppressed. The conventional heatmap loss function can be regarded as our special case where $\mu = 0$.  

\vspace{-0.1in}
\subsubsection{Discussion}
A conventional SPPE depends on a high-quality human detection result. Its tasks are locating and identifying target joints according to the given human proposal. If SPPE mistakes interference joints for target joints, it will be an unrecoverable error. Missing joints cannot be restored in the post-processing step like pose-NMS.

Our proposed joints candidate loss is aimed to tackle this limitation. This loss function encourages JC SPPE network to predict multi-peak heatmaps and sets all the possible joints as candidates. In crowded scenes, while conventional SPPE is hard to identify target joints, JC SPPE can still predict a list of joint candidates and guarantee high recall. We leave the association problem to the next procedure, where we have more global information from other JC SPPEs (on other human proposals) to solve it.

\subsection{Person-Joint Graph} \label{sec:graph}
Due to our joint-candidate mechanism and redundant human proposals from human detector, joint candidates are numerically much greater than the actual joint numbers. To reduce redundant joints, we build a person-joint graph and apply a maximum person-joint matching algorithm to construct the final human poses. 
\begin{figure}[htb]
\begin{center}
\includegraphics[width=1\linewidth]{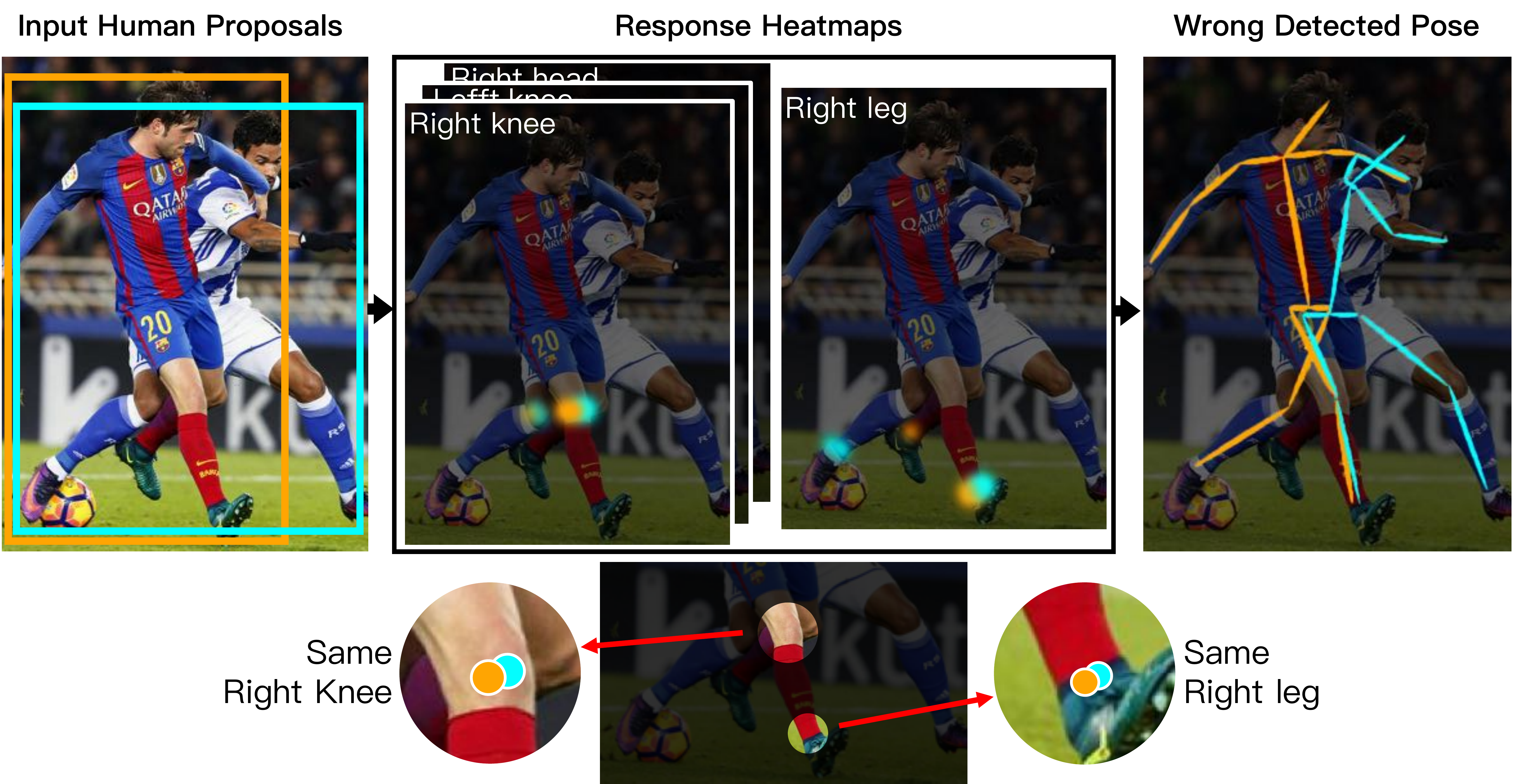}
\end{center}
   \caption{In crowded scenes, human proposals are highly overlapped. Overlapped human proposals tend to predict same actual joint. In this example, if we directly connect the highest response to build final poses, two human proposals will locate same right knee and right leg. Our proposed association algorithm can solve this problem by globally best matching.}
\label{fig:exm}
\end{figure}

\vspace{-0.2in}
\subsubsection{Joint Node Building}\label{sec:joint_node} Since highly overlapped human proposals tend to predict the same actual joint (as Figure~\ref{fig:exm}), we first group these candidates that represent the same actual joint as one joint node.

Thanks to the high-quality joint prediction, candidate joints that indicate the same joint are always close to each other. Thus, we can group them using the following criterion: given two candidate joints located at $p_1^k$ and $p_2^k$ with control deviation $\delta^k$, we label them as the same group, if
\begin{equation} 
    ||p_1^{(k)} - p_2^{(k)}||_2 \leq  \min \{u_1^k, u_2^k\} \delta^{(k)},  
\end{equation}
where $u_1^k$ and $u_2^k$ are the Gaussian response size of two joints on heatmaps, determined by the Gaussian response deviation. $\delta^{(k)}$ is the parameter for controlling deviation of the $k^{th}$ joint, which we directly adopt from MSCOCO keypoint dataset~\cite{mscoco}. The reason why we use $\min \{u_1, u_2\}$ rather than a constant threshold is to guarantee that, only if $p_1$ and $p_2$ fall into each others' control domain (radii are $u_1^k \delta^{k}$, $u_2^k \delta^{k}$) simultaneously, we group them together. One node represents a group of joints that cluster together by the above criterion.

Now, by building a joint group as one node, we have joint node set $\mathcal{J} = \{v_j^{k} : \text{for}\ k \in \{1, \dots, K\}, j \in \{1, \dots, N_k\}\}$, where $N_k$ is the number of joint nodes of body part $k$, $v_j^{k}$ is the $j^{th}$ node of body part $k$. The total number of joint nodes in $\mathcal{J}$ is $\sum_{k} N_k$.

\vspace{-0.1in}
\subsubsection{Person Node Building}
Person nodes represent the human proposals detected by human detector. We denote person node set as $\mathcal{H} = \{h_i : \forall i \in \{1 \dots M\}\}$, where $h_i$ is the $i^{th}$ person node, and $M$ is the number of detected human proposals.

Ideally, a qualified human proposal tightly bounds a human instance. However, in crowded scenes, this condition is not always satisfied. The human detector will produce many redundant proposals, including truncated and incompact bounding boxes. We will eliminate these low-quality person nodes during global person-joint matching in Sec.~\ref{sec:optimize}.

\vspace{-0.1in}
\subsubsection{Person-Joint Edge} After obtaining the node of both joints and persons, we connect them to construct our person-joint graph. If a joint node $v_j^{k}$ contains a candidate joint from person node $h_i$, we build an edge $e_{i,j}^k$ between them. The weight of $e_{i,j}^k$ is the response score of that candidate joint, which is denoted as $w_{i,j}^k$. In this way, we can construct edge set $\mathcal{E} = \{e_{i,j}^k: \forall i,j,k\}$.

The person-joints graph can then be written as:
\begin{equation}
    \mathcal{G} = ((\mathcal{H}, \mathcal{J}), \mathcal{E}).
\end{equation}

\subsection{Globally Optimizing Association} \label{sec:optimize}
From now on, our goal of estimating human poses in the crowd is transformed into solving the above person-joint graph and maximizing the total edge weights. We have our objective function as:
\begin{align}
    \max_d \mathcal{G} = &\max_{d} \sum_{i, j, k} w_{i,j}^{(k)} \cdot d_{i,j}^{(k)} \\
    s.t. \quad &\sum_{j} d_{i,j}^{(k)} \leq 1, \quad
    \begin{aligned}
        &\forall k \in \{1, \dots, K\},\\
        &\forall i \in \{1, \dots, M\}
    \end{aligned} \label{st:st1}\\
    &\sum_{i} d_{i,j}^{(k)} \leq 1, \quad
    \begin{aligned}
        &\forall k \in \{1, \dots, K\},\\
        &\forall j \in \{1, \dots, N_k\}
    \end{aligned} \label{st:st2} \\
    &d_{i,j}^{(k)} \in \{0, 1\}, \quad \forall i,j,k \label{st:st3}
\end{align}
where $d_{i,j}^{(k)}$ indicates whether we keep the edge $e_{i,j}^k$ in our final graph or not. The constraints of Eq.~\ref{st:st1} and \ref{st:st2} enforce that each human proposal can only match at most one $k^{th}$ joint.

Note that $\mathcal{G}$ can be decomposed into $K$ sub-graph $\mathcal{G}_k = ((\mathcal{H}, \mathcal{J}^{(k)}), \mathcal{E}^{(k)})$, where
$\mathcal{J}^k = \{v_j^{(k)} : \forall j \in \{1 \dots N_k\}\}$ and $\mathcal{E}^k = \{e_{i,j}^{(k)} : \forall i \in \{1 \dots M\}, j \in \{1 \dots N_k\} \}$.
Thus, our objective function can be formulated as
\begin{align}
    \max_d\mathcal{G} &= \max_{d} \sum_{i, j, k} w_{i,j}^{(k)} \cdot d_{i,j}^{(k)} \\
    &= \sum_{k=1}^K (\max_{d^{(k)}} \sum_{i, j} w_{i,j}^{(k)} \cdot d_{i,j}^{(k)})\\
    &= \sum_{k=1}^K \max_{d^{(k)}} \mathcal{G}_{k}.
    \label{eq:degrade}
\end{align}

As shown in Eq.~\ref{eq:degrade}, solving the global assignment problem in person-joint graph $\mathcal{G}$ is mathematically equivalent to solving its sub-graph $\mathcal{G}_k$ separately. $\mathcal{G}_k$ is a bipartite graph that composed of person subset and the $k^{th}$ joint subset. For each sub-graph, the updated Kuhn-Munkres algorithm~\cite{complexity} is applied to get the optimized result. By addressing each $\mathcal{G}_k$ respectively, we obtain the final result set $\mathcal{R}$.

Given the graph matching result, if $d_{i,j}^{(k)} = 1$ the weighted center of $v_{j}^k$ is assigned to the $i^{th}$ human proposal as its $k^{th}$ joint. Here, weighted center means the linear combination of candidate joints coordinate in $v_{j}^k$ and the weights are their heatmap response scores. In this way, the pose of each human proposal can be constructed. The person nodes that can not match any joint will be removed.

\vspace{-0.1in}
\paragraph{Computational Complexity}
The inference speed of pose estimation is essential in many applications. We prove that our global association algorithm is as efficient as common greedy NMS algorithms.

As the hereditary property identified by White and Whiteley~\cite{klsparse}, a graph $G$ is $(k,l)-sparse$ if every nonempty sub-graph $X$ has at most $k|X| - l$ edges, where $|X|$ is the number of vertices in sub-graph $X$ and $0 \leq l < 2k$.

\begin{figure}[htb]
\begin{center}
\includegraphics[width=0.9\linewidth]{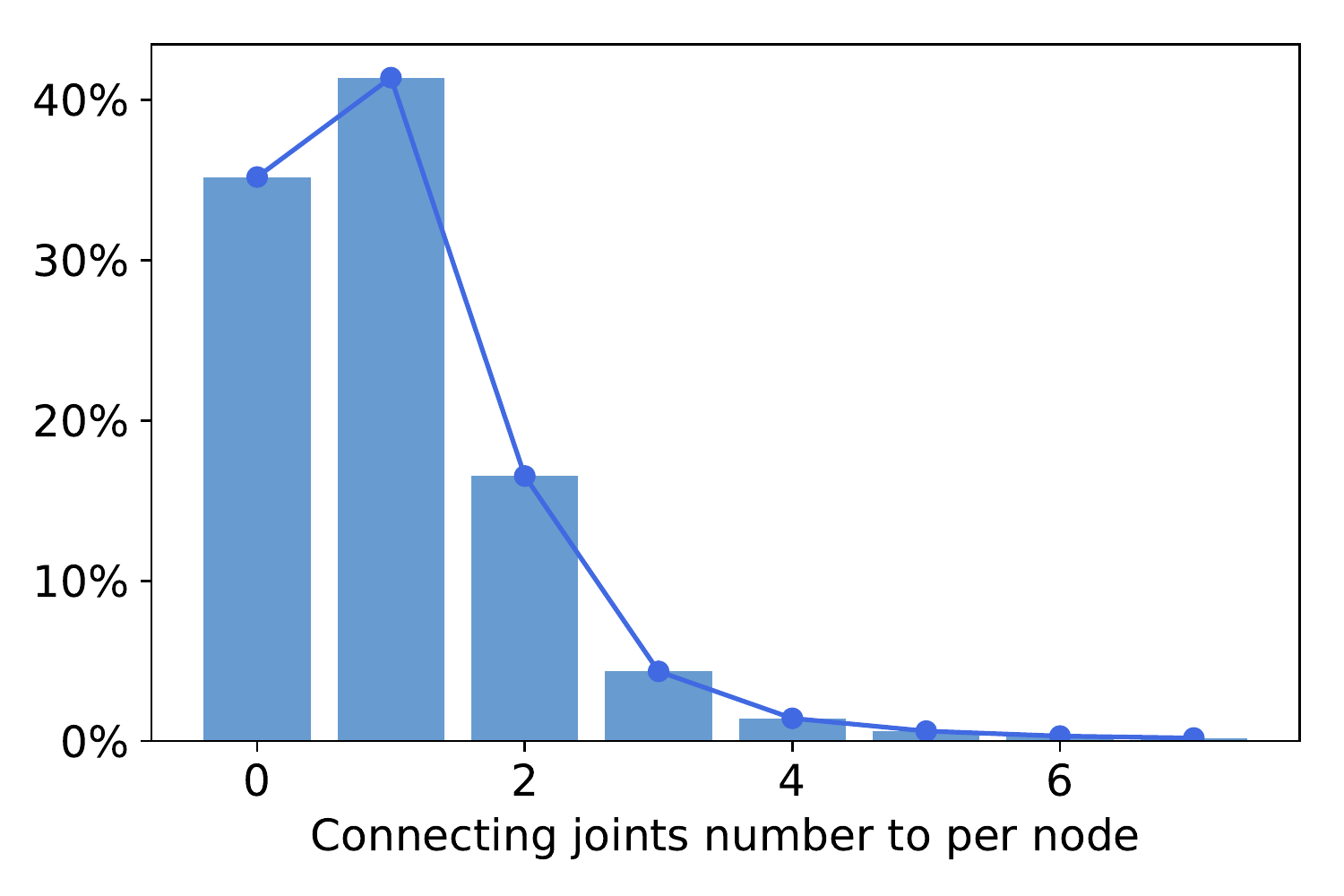}
\end{center}
   \caption{Instance-Joint connection distribution. The x-axis denote the number of human bounding boxes that cover a same joint. This statistical result is based on the ground truth annotations.}
\label{fig:cruve}
\vspace{-0.15in}
\end{figure}

Consider the sub-graph $\mathcal{G}^{(k)} = ((\mathcal{H}, \mathcal{J}^{(k)}), \mathcal{E}^{(k)})$. It represents the connection between human proposals and the $k^{th}$ type of joints. According to our statistics (Fig.~\ref{fig:cruve}), every human bounding box covers four persons at most in crowded scenes. Therefore, one person node builds connection edges to $4$ joints at most. In other words, our person-joint sub-graph $\mathcal{G}^{(k)}$ is $(4, 0)-sparse$ since
\begin{equation}
    |\mathcal{E}^{(k)}| \leq 4|\mathcal{G}^{(k)}| - 0.
\end{equation}

Due to the sparsity of our person-joint graph, we can solve the association problem efficiently. We transform $\mathcal{E}^{(k)}$ into an adjacency matrix $M_{e^k}$ (unconnected nodes refer to $0$). According to the work of Carpaneto \etal~\cite{complexity}, this linear assignment problem for the sparse matrix can be solved in $\mathcal{O}(n^2)$, i.e., $\mathcal{O}((|\mathcal{H}| + |\mathcal{J}^{(k)}|)^2)$. Since we have eliminated the redundant joints and there is a one-to-one correspondence between joints and persons, the expectation of $|\mathcal{J}^{(k)}|$ is equal to $|\mathcal{H}|$. Thus we have $\mathcal{O}((|\mathcal{H}| + |\mathcal{J}^{(k)}|)^2) = \mathcal{O}(|\mathcal{H}|^2)$. Such computation complexity is the same as the complexity of conventional greedy NMS algorithms.

\subsection{Discussion} Our method adopts the graph-based approach to associate joints with human proposals in a globally optimal manner. Human proposals compete with each other for joint nodes. In this way, unqualified human proposals without dominant human instance would fail to be assigned any joints, since their joint response scores are all relatively low due to missing dominant human instance. Therefore, many redundant and poor human proposals are rejected. In comparison to our approach, conventional NMS is a greedy and instance-based algorithm, which is less effective. Although~\cite{nms1, nms2, alphapose} proposed pose-NMS to utilize pose information, their algorithms are based on instances and cannot tackle the missing joints and wrong assembling problem. Our globally optimizing association method can deal with such situations well.

\section{CrowdPose Dataset}
In this section, we introduce another contribution of our paper, namely, CrowdPose dataset, including crowded scenes definition, data collection process, and the dataset statistics.

\subsection{Crowding Level Definition}
To build a dataset of crowded human pose, we need to define a \textit{Crowd Index} first, which measure the crowding level in a given image.

Intuitively, the number of persons in an image seems to be a good measurement. However, the principal obstacle to solving crowded cases is not caused by the number of persons, but rather by occlusion in a crowd. Therefore, we need a new \textit{Crowd Index} to indicate crowding level. In the bounding box of the $i^{th}$ human instance, we denote the number of joints that belonging to the $i^{th}$ person and other (not $i^{th}$) persons as $N_i^a$ and $N_i^b$ respectively. $N_i^b/N_i^a$ is the \textit{crowd ratio} of the $i^{th}$ human instance. Our \textit{Crowd Index} is derived by averaging the crowd ratio of all persons in an image:
\begin{equation}
    Crowd~Index = \frac{1}{n}\sum_{i=1}^{n} \frac{N_i^b }{N_i^a},
\end{equation}
where $n$ indicates the total number of persons in the image.

\begin{figure}[tb]
\centering
\subfigure[MSCOCO]{\label{fig:crowd_distribute:a}
\includegraphics[width=0.45\linewidth]{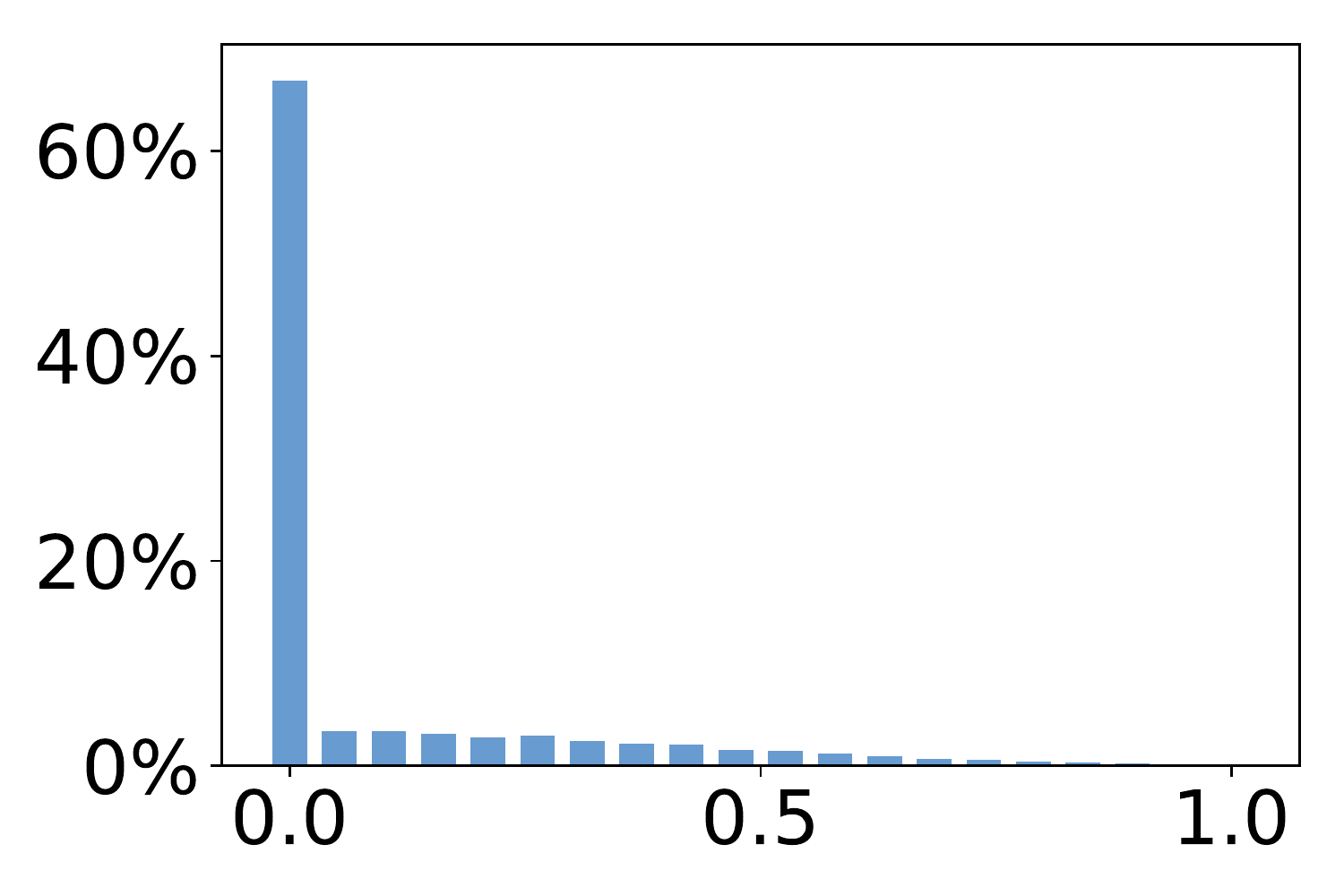}}
\hspace{0.01\linewidth}
\subfigure[MPII]{\label{fig:crowd_distribute:b}

\includegraphics[width=0.45\linewidth]{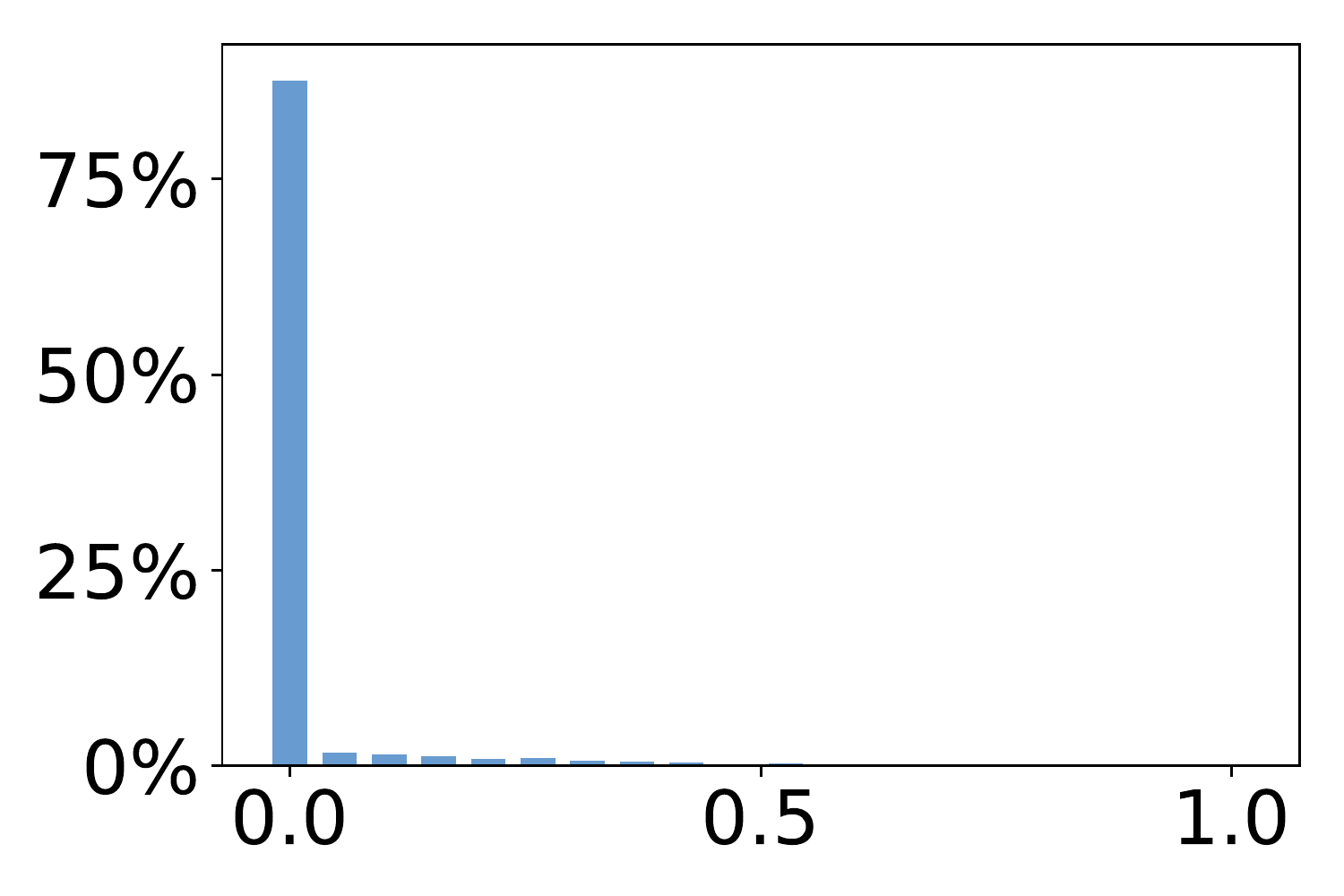}}
\vfill
 
\subfigure[AI Challenger]{\label{fig:crowd_distribute:c}
\includegraphics[width=0.45\linewidth]{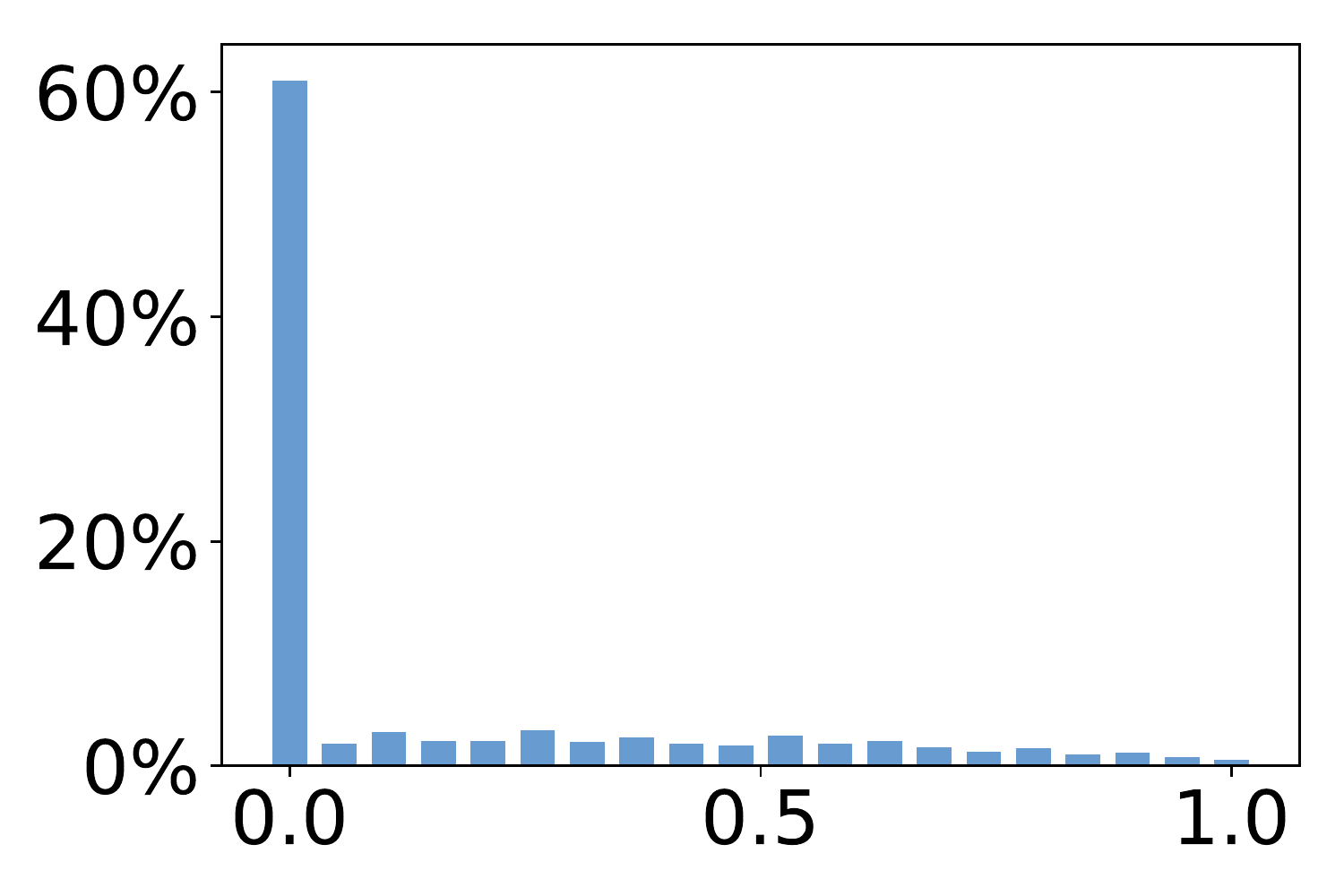}}
\hspace{0.01\linewidth}
\subfigure[CrowdPose]{\label{fig:crowd_distribute:d}
\includegraphics[width=0.45\linewidth]{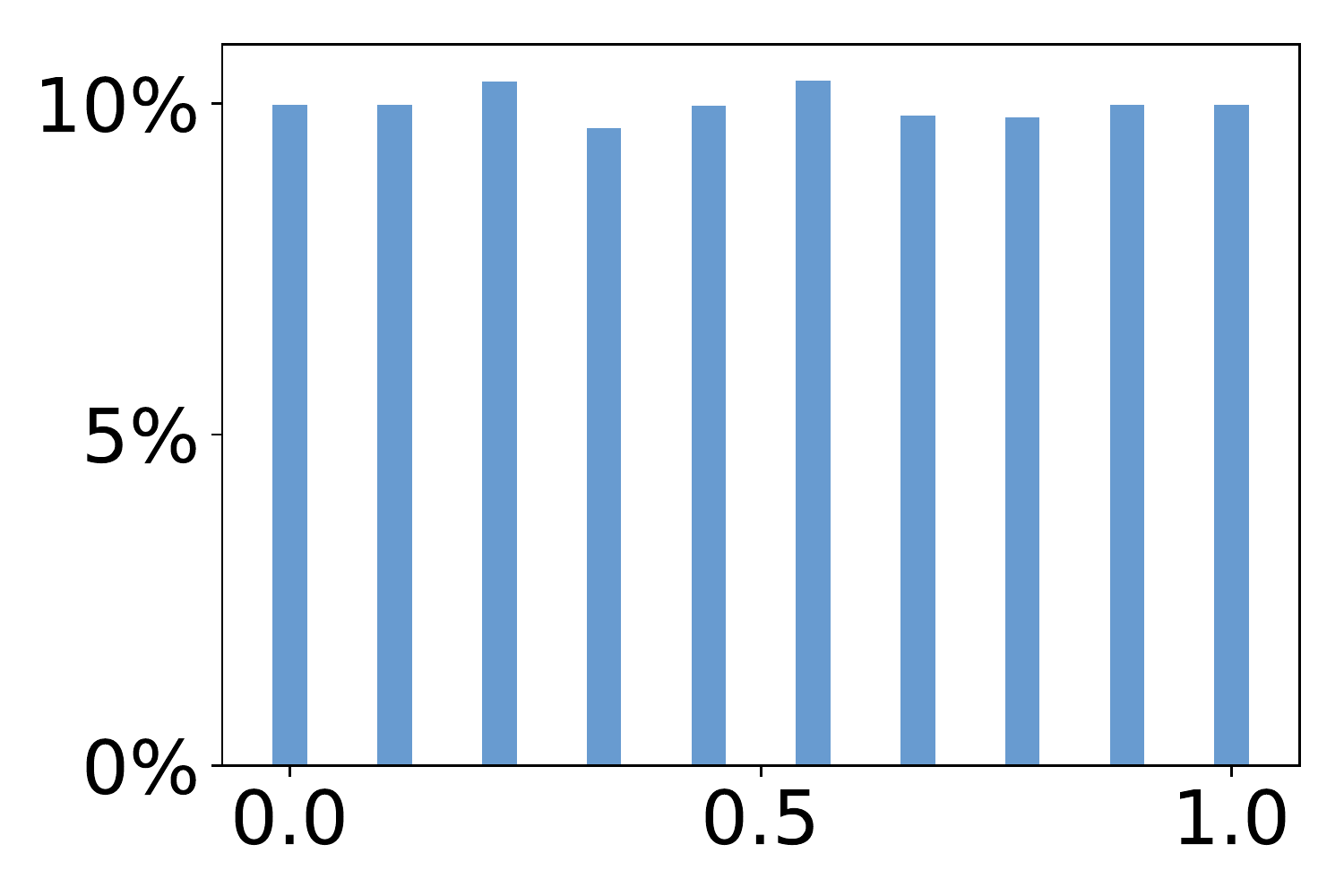}}
\caption{$Crowd Index$ distributions of current popular datasets and our dataset. Three public datasets are dominated by uncrowded images. Meanwhile, our CrowdPose dataset has a near uniform distribution.}

\label{fig:crowd_distribute}
\end{figure}

We evaluate the \textit{Crowd Index} distribution of three public benchmarks: MSCOCO (person subset), MPII and AI Challenger. As shown in Figure~\ref{fig:crowd_distribute}, uncrowded scenes dominate these benchmarks, which leads the state-of-the-art methods only focus on these simple cases and ignore the crowded ones.


\begin{figure*}[!t]
\begin{center}
\includegraphics[width=1\linewidth]{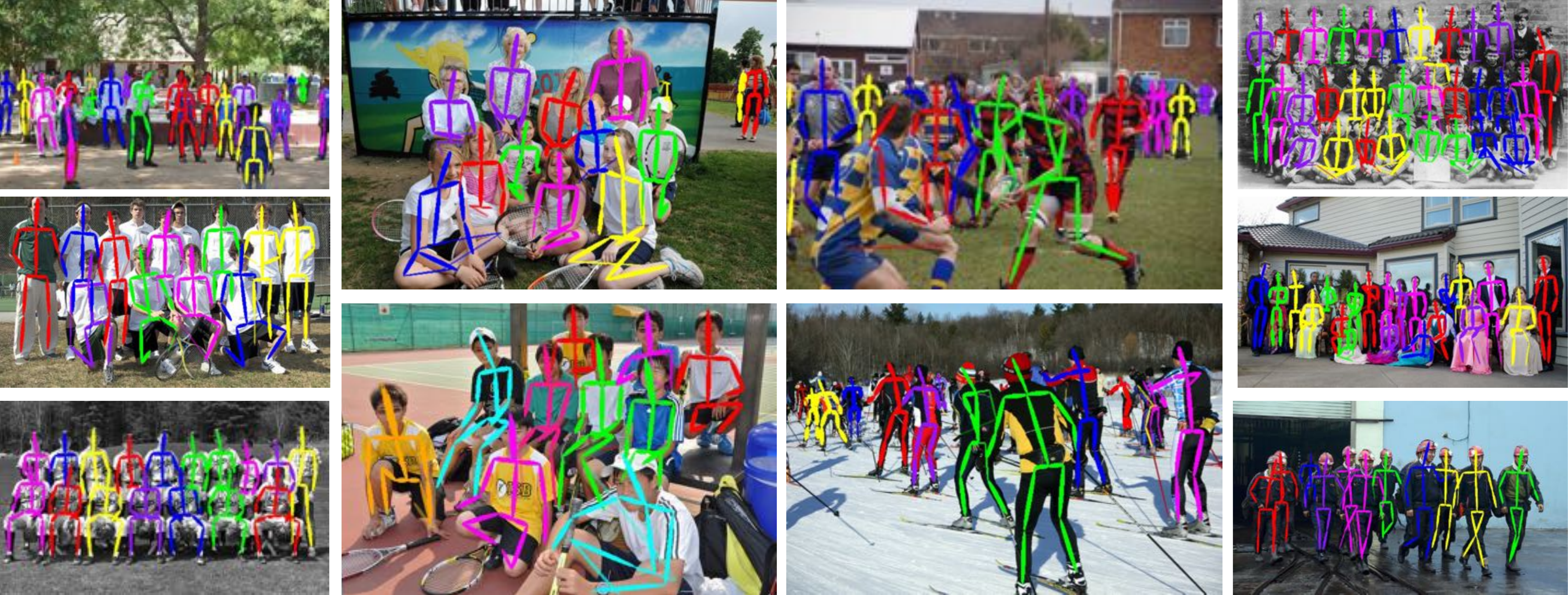}
\end{center}
   \caption{Qualitative results of our model’s predictions is presented. Different person poses are painted in different colors to achieve better visualization.}
\label{fig:res}
\end{figure*}

\begin{table*}
\begin{center}
\begin{tabular}{|c|c|c|c|c|c|c|}
\hline
Method & mAP @0.5:0.95 & mAP@0.5 & mAP @0.75 & mAR @0.5:0.95 & mAR @0.5 & mAR @0.75\\
\hline\hline

Mask R-CNN~\cite{maskrcnn} & 57.2 & 83.5 & 60.3 & 65.9 & 89.5 & 69.4\\
AlphaPose~\cite{alphapose} & 61.0 & 81.3 & 66.0 & 67.6 & 86.7 & 71.8\\
Xiao \etal~\cite{msra} & 60.8 & 81.4 & 65.7 & 67.3 & 86.3 & 71.8\\
\hline
\textbf{Ours} & \textbf{66.0} & 84.2 & 71.5 & \textbf{72.7} & 89.5 & 77.5\\
\hline

\end{tabular}
\end{center}
\caption{Results on CrowdPose test set.} \label{tb:method}
\end{table*}

\subsection{Data Collection}
To set up a benchmark that covers various scenes and encourages models to adapt to different kinds of situations, we wish our benchmark covers not only crowded cases but also simple daily life scenes. To achieve that, we first analyze three public benchmarks~\cite{mscoco, mpii, aic} and divide their images into 20 groups according to \textit{Crowd Index}, ranging from 0 to 1. The step among different groups is 0.05. Then we sample 30,000 images uniformly from these groups in total.

\subsection{Image Annotation}
Although these images have been annotated, their label formats are not fully aligned. In terms of annotated joints number, MSCOCO has 17 keypoints, while MPII has 16 and AI Challenger annotates 14 keypoints. Meanwhile, human annotators are easier to make mistakes in the crowded cases. Thus, we re-annotate these images by the following steps.
\begin{itemize}
    \item We use 14 keypoints definition and annotate keypoints and full-body bounding boxes for persons in 30,000 images.
    \item We analyze the \textit{Crowd Index} for 30,000 images again with new annotations, and select 20,000 high-quality images.
    \item We further crop each person in the images, and then annotate the interference keypoints in each bounding box.
\end{itemize}

We use cross annotation, which means at least two annotators annotate each image. If these two annotations have a large deviation, we regard them as mistakes and re-annotate this image. Finally, we take the average value of each keypoint location to ensure the annotation quality.

\subsection{Dataset Statistics}
\paragraph{Dataset Size} In total, our dataset consists of 20,000 images, containing about 80,000 persons. The training, validation and testing subset are split in proportional to 5:1:4.

\vspace{-0.1in}
\paragraph{Crowd Index Distribution}
The \textit{Crowd Index} distribution of CrowdPose is shown in Figure~\ref{fig:crowd_distribute} (d). Unlike the other datasets, CrowdPose has a uniform distribution of \textit{Crowd Index}. Note that we do not simply force CrowdPose to reach high \textit{Crowd Index}. If a model is only trained on crowded scenes, it may degrade its performance on uncrowded cases due to the bias of training set. Uniform distribution can promote a model to adapt to various scenes.

\paragraph{Average IoU}
We further calculate the average intersection over union(IoU) of human bounding boxes. It turns out that CrowdPose has an average bounding box IoU of 0.27, while MSCOCO, MPII, and AI Challenger have 0.06, 0.11 and 0.12 respectively.
\begin{table}
\begin{center}
\small
\begin{tabular}{|c|ccc|c|}
\hline
Method & $\text{AP}_{Easy}$ & $\text{AP}_{Medium}$ & $\text{AP}_{Hard}$ & FPS\\
\hline\hline
OpenPose~\cite{openpose} & 62.7 & 48.7 & 32.3 & 5.3 \\
Mask R-CNN~\cite{maskrcnn} & 69.4 & 57.9 & 45.8 & 2.9\\
AlphaPose~\cite{alphapose} & 71.2 & 61.4 & 51.1 & 10.9\\
Xiao \etal~\cite{msra} & 71.4 & 61.2 & 51.2 & - \\
\hline
\textbf{Ours} & \textbf{75.5} & \textbf{66.3} & \textbf{57.4} & 10.1\\
\hline

\end{tabular}
\end{center}
\caption{Results on CrowdPose test set. Test set is divided into three part and we report the results respectively. FPS column reports the runtime speed on the whole test set.} \label{tb:method2}
\end{table}

\section{Experiments}
In this section, we first introduce datasets and settings for evaluation. Then we report our results and comparisons with state-of-the-art methods, and finally conduct ablation studies on components in our method.

\subsection{Datasets}
\paragraph{CrowdPose} Our proposed CrowdPose dataset contains 20,000 images in total and 80,000 human instances. Its \textit{Crowd Index} satisfies uniform distribution in $[0, 1]$. CrowdPose dataset aims to promote performance in crowded cases and make models generalize to different scenarios.

\paragraph{MSCOCO Keypoints} We also evaluate our method on the MSCOCO Keypoints dataset~\cite{mscoco}. It contains over 150,000 instances for training and 80,000 instances for testing. The persons in this dataset overlap less frequently than CrowdPose, and it has a \textit{Crowd Index} centralized near zero.

\subsection{Evaluation Metric}
We follow the evaluation metric of MSCOCO, using average precision (AP) and average recall (AR) to evaluate the result. Object keypoint similarity (OKS) plays the same role as the IoU to adopt AP/AR for keypoints detection. We consider \textbf{mAP}, averaged over multiple OKS values (.50:.05:.95), as our primary metric. Moreover, we divide the CrowdPose dataset into three crowding levels by \textit{Crowd Index}: easy (0-0.1), medium (0.1-0.8) and hard (0.8-1), to better evaluate our model performance in different crowded scenarios. We use the same keypoint standard deviations as MSCOCO when calculating OKS in all experiments. 

\subsection{Implementation Details}

Our method follows the two-step framework. Since human detector and pose estimation network are not what we focus on, we simply adopt the human detector (YoloV3~\cite{yolov3}) and pose estimation network provided by AlphaPose~\cite{alphapose} which is a state-of-the-art two-step method. During the training step, we adopt rotation ($\pm30$), scaling ($\pm30\%$) and flipping data augmentation. The input resolution is $320 \times 256$ and the output heatmap resolution is $80 \times 64$. The learning rate is set to $1 \times 10^{-4}$ and $1 \times 10^{-5}$ after 80 epochs. Mini-batch size is set to 64, and RMSprop~\cite{rmsprop} optimizer is used. During testing, the detected human bounding boxes are first extended by $30\%$ along both the height and width directions and then forwarded through the Joint-candidate SPPE. The locations of joint candidates are obtained from the averaged output heatmaps of original and flipped input image. We conduct our experiments on two Nvidia 1080Ti GPUs. Our whole framework is implemented in PyTorch.

For comparison with current state-of-the-art methods~\cite{maskrcnn, msra, alphapose} on CrowdPose dataset, we retrain them based on the configuration provided by the authors. To be fair, we use ResNet-101 as backbone for all SPPE networks and use the same human detector for~\cite{msra}. For Mask R-CNN we also use the FPN-based ResNet-101 backbone. Same training batch size is used for a fair comparison.

\begin{table}
\begin{center}
\small
\begin{tabular}{|c|c|c|c|c|c|c|}
\hline
Method & mAP @0.5:0.95 & mAR @0.5:0.95\\
\hline\hline
Mask R-CNN~\cite{maskrcnn} & 64.8 & 71.1 \\
AlphaPose~\cite{alphapose} & 70.1 & 74.4\\
Xiao \etal~\cite{msra} & 69.8 & 74.1\\
\hline
\textbf{Ours} & \textbf{70.9} & \textbf{76.4}\\
\hline

\end{tabular}
\end{center}
\caption{Results on MSCOCO test-dev set. We compare the state-of-the-art methods with same detection backbone.} \label{tb:method3}
\end{table}
\begin{table}
\begin{center}
\small
\begin{tabular}{|cc|c|c|c|}
\hline
&Method & mAP & mAR\\
\hline\hline

&\textbf{Ours} ($\text{SPPE}^{+}$+Association) & \textbf{66.0} & \textbf{72.7} \\
\hline
(a)&w/o joint-candidate Loss & 61.7 & 68.7 \\
\multirow{2}{*}{(b)}&Greedy NMS & 49.1 & 63.1 \\
&Parametric Pose-NMS~\cite{alphapose} & 64.2 & 71.1 \\
\hline
\end{tabular}
\end{center}
\caption{Results on CrowdPose test set. mAP and mAR are the average value over multiple OKS values
(0.50:0.05:0.95). ``w/o X'' means without X module in our pipeline. ``Greedy NMS'' means a conventional NMS method based on proposal scores and IoU.} \label{tb:ablation}
\end{table}

\subsection{Results}
\paragraph{CrowdPose}
Quantitative results on CrowdPose test set are given in Table~\ref{tb:method}. Our method achieves \textbf{5.2} mAP higher than state-of-the-art methods.  It demonstrates the effectiveness of our proposed method to tackle the problem of pose estimation in crowded scenes. To further evaluate our method in crowded scenes, we report the results on three crowding level in Table~\ref{tb:method2}, \ie, uncrowded, medium crowded and extremely crowded. Notably, our method improves \textbf{4.1} mAP in uncrowded scenes, while achieves \textbf{4.9} and \textbf{6.2} mAP higher in medium crowded and extremely crowded scenes separately. This result demonstrates that our method has superior performance in crowded scenes. We present some qualitative results in Figure~\ref{fig:res}. More results will be given in supplementary file.

\vspace{-0.1in}
\paragraph{MSCOCO}
We also evaluate our method on MSCOCO dataset to show the generalization ability of our method and results are given in Table~\ref{tb:method3}. Without bells and whistles, our method achieves 70.9  mAP on COCO test-dev set. It brings \textbf{0.8} mAP improvements over AlphaPose~\cite{alphapose} given the same human detector and SPPE network, which proves that our method can perform general improvement on pose estimation problem. Note that for~\cite{msra}, the results reported in their paper use a strong human detector which is not open-sourced. To make a fair comparison, we report the results using YOLOV3 as human detector.

\vspace{-0.1in}
\paragraph{Inference Speed}
The runtime speed of fully open-sourced method is tested and shown in Table~\ref{tb:method2}. We obtain the FPS results by averaging the inference time on the test set. To achieve the best performance, we use the most accurate configuration for OpenPose~\cite{openpose}, which has an input resolution of $1024 \times 736$. As shown in the table, our method achieves 10.1 FPS on the test set, which is slightly slower than AlphaPose~\cite{alphapose} but faster than other methods. It proves that our method works the most accurate yet very efficient in crowded cases.

\subsection{Ablation Studies}
We study different components of our method on CrowdPose test set, as reported in Table \ref{tb:ablation}.

\vspace{-0.15in}
\paragraph{Joints Candidate Loss}
We first evaluate the effectiveness of our joint-candidate loss. In this experiment, we replace our joint-candidate loss with mean square loss, which is widely used in state-of-the-art pose estimation methods. The experimental result is shown in Table \ref{tb:ablation}(a). The final mAP drops from 66.0\% to 61.7\%. It proves that our loss function can encourage SPPE to predict more possible joints and resist interference.

\vspace{-0.1in}
\paragraph{Globally Optimizing Association}
Next, we compare our association algorithm to several NMS algorithms, including bounding box NMS, poseNMS~\cite{nms1, nms2} and parametric poseNMS~\cite{alphapose}. The experimental results are shown in Table \ref{tb:ablation}(b). We can see that our association algorithm greatly outperforms previous methods. We note that these NMS algorithms are all instance-based. They eliminate redundancy on instance level. Instance-based elimination is not the best solution for pose estimation problem. Because of the complexity of human pose, we need to reduce redundancy on joints level. Meanwhile, all the NMS algorithms are greedy algorithms essentially. Therefore, their results may be not the global optimum. Our association algorithm can give the global optimal result while running as efficient as previous NMS algorithm.

\vspace{-0.1in}
\section{Conclusion}
In this paper, we propose a novel method to tackle the occlusion problem of pose estimation. By building a person-joint graph based on the outputs of our joint candidates SPPE, we transform the pose estimation problem into a graph matching problem and optimize the results in a global manner. To better evaluate the model performance in crowded scenes, we set up a CrowdPose dataset with a normal distribution of \textit{Crowd Index}. Our proposed method significantly outperforms the state-of-the-art methods in CrowdPose dataset. Experiments on MSCOCO dataset also demonstrate that our method can generalize to different scenarios.

{\small
\bibliographystyle{ieee}
\bibliography{egbib}
}
\end{document}